\documentclass{article}
\usepackage{spconf,amsmath,graphicx}

\title{modality attention for end-to-end audio-visual speech recognition}
\name{Pan Zhou$^{1}$, Wenwen Yang$^{2}$, Wei Chen$^{2}$, Yanfeng Wang$^{2}$, Jia Jia$^{1}$
}
\address{$^{1}$Department of Computer Science and Technology, Tsinghua University, Beijing, P.R.China
\\ $^{2}$Voice Interaction Technology Center, Sogou Inc., Beijing, P.R.China
\\\small \texttt{\{zh-pan,jjia\}@mail.tsinghua.edu.cn,}\;\texttt{\{yangwenwen,chenweibj8871,wangyanfeng\}@sogou-inc.com}
}
\begin{document}
\ninept
\maketitle
\begin{abstract}
Audio-visual speech recognition (AVSR) system is thought to be one of the most promising solutions for robust speech recognition, especially in noisy environment. In this paper, we propose a novel multimodal attention based method for audio-visual speech recognition which could automatically learn the fused representation from both modalities based on their importance. 
Our method is realized using state-of-the-art sequence-to-sequence (Seq2seq) architectures.
Experimental results show that  relative improvements from 2\% up to 36\% over the auditory modality alone are obtained depending on the different signal-to-noise-ratio (SNR). Compared to the traditional feature concatenation methods, our proposed approach can achieve better recognition performance under both clean and noisy conditions. We believe modality attention based end-to-end method can be easily generalized to other multimodal tasks with correlated information.
\end{abstract}
\section{Introduction}
\label{sec:intro}

Artificial intelligence attracts more and more attention in recent years. Man-machine interaction interface for smart devices is necessary. Speech is the most natural and convenient way of communication between people. Consequently, automatic speech recognition (ASR) is considered to be the first choice for effective man-machine interaction. In tasks of some quiet environment, recent state-of-the-art ASR systems \cite{chiu2017state-of-art-attention} are capable of reaching a recognition accuracy above 95\%.  It is recognized that humans understand speech not only by listening but also by considering visual cues of lips and face \cite{mcgurk1976hearinglips}. So speech interaction is multimodal in nature although we don't rely much on visual information.  This is probably because auditory modality contains most of the useful information that is enough for our understanding of other people. The visual modality becomes important when the acoustic signal is distorted by noise.  AVSR aims to use visual information derived from lip motions to complement corrupted audio speech signal.

Traditional audio based ASR system is made up of acoustic model (AM), language model (LM), pronunciation model and decoder. These ASR systems become more and more accurate after hidden Markov model (HMM) had been used for speech recognition. This progress can be attribute to not only  the appropriate acoustic feature, such as Mel-frequency cepstral coefficients (MFCC) and filter banks, but also the more expressive model like Gaussian mixture model (GMM) and deep neural networks (DNN). Recurrent neural network (RNN), such as long short term memory (LSTM) handles temporal dependency by different gates, which makes it more appropriate for acoustic modeling. Although DNN and LSTM boost speech recognition accuracy a lot, ASR is still a complex system and need to tune each component separately. There is a favor in end-to-end ASR approach recent years which tries to combine all components into a single neural network and optimize them jointly. Connectionist temporal classification (CTC) \cite{graves2006ctc, graves2014ctc}, RNN transducer (RNN-T) \cite{graves2012rnnt, graves2013ctcrnnt, rao2017googleRNNT} and attention based encoder-decoder \cite{bahdanau2014neural,chorowski2014endFirst, chan2016las} are the three main approaches. Such end-to-end systems alleviate effect of HMM and output meaningful units like phones and characters directly. The work of \cite{chiu2017state-of-art-attention} shows that attention based listen, attend and spell (LAS) model can exceed traditional NN-HMM hybrid systems in speech tasks like dictation and voice search. 

Understanding speech from visual signal alone, i.e. lipreading, has been of interest for decades. Many researchers dedicated to extract powerful visual features for precise lipreading models. These feature extraction approaches include top-down approach and bottom-up approach. The former one uses priori lip shape representation within a model, such as active shape models (ASMs) \cite{luettin1996ASM} and active appearance models (AAMs) \cite{cootes2001AAM}, and extracts model-based features. The latter one estimates visual features from images directly by discrete cosine transform (DCT), principal component analysis (PCA) and discrete wavelet transform (DWT) etc \cite{matthews2001comparison,aleksic2004comparison}. Thanks to the development in neural networks recent years, learning visual feature representation automatically by supervised training of convolutional neural networks (CNN) or LSTM has improved lipreading significantly \cite{noda2014lipreadingCNN,tatulli2017featureCNN,chung2016lipreadWild,wand2016lipreadingLSTM}. After designing or learning visual features, GMM-HMM based model or softmax classification model is trained to predict isolated words or phonemes been spoken. Just like the development of ASR, LipNet \cite{assael2016lipnet} utilizes CNN and LSTM with CTC loss to train a neural net entirely end-to-end and operates sentence level lipreading. Another representative work is \cite{chung2017lip} which proposed watch, attend and spell (WAS) to transcribe large vocabulary visual speech at the character level.
 
The work of \cite{petajan1984lipreading} proposed to use lipreading to enhance ASR. Integrating both auditory and visual information simultaneously into a single model for AVSR has been a challenge. One reason is lipreading is indeed a more difficult problem because human lip movements carry less information than speech. Another reason is the demand for an effective fusion strategy of two correlated but inherently different data streams. Some of the earlier audio-visual fusion methods are reviewed in \cite{katsaggelos2015audiovisualfusion, potamianos2003recent}. Noda \cite{noda2015audiovisual} makes use of CNN to extract visual feature and combines it with audio feature by a multi-stream HMM. Feature fusion and then using a one stream model is the dominant approach \cite{petridis2018end,chung2017lip, sterpu2018AValign} in AVSR.  \cite{sterpu2018AValign} correlates every frame of acoustic feature with visual context feature acquired by cross modality attention. The correlated feature is then decoded by an attention based decoder. The watch, listen, attend and spell (WLAS) model \cite{chung2017lip} concatenates two context vectors obtained by attending over individual modality to predict the output units. However, as pointed out by \cite{potamianos2003recent} concatenation of features does not explicitly make model to learn stream reliabilities. Furthermore, WLAS is observed to become over-reliant on the audio modality and is hard to train well. Consequently, a regularization technique where one of the streams is randomly dropped during training is applied. 

In this work, we propose a novel multi-modality attention method to integrate information from audio and video for audio-visual speech recognition. The key contribution of our approach is that we use an additional attention mechanism that enables the model to automatically adjust its modality attention to a more reliable input modality. Furthermore, by using an LSTM to model temporal variability for each modality, the attention of different modality can change over time. We evaluate the use of modality attention mechanism in WLAS modal architecture. Experimental results demonstrate that our modality attention equipped end-to-end system can outperform the audio-only system up to 36\% relative improvement in 0dB SNR, showing the effectiveness of our proposal.

The remainder of the paper is structured as follows. We present the multimodal attention approach in details in Section \ref{sec:modalAttention}. Experimental results are given in Section \ref{sec:exps} and the paper is concluded with our findings and future work in Section \ref{sec:conclusion}.

\section{End-to-end multimodal attention AVSR}
\label{sec:modalAttention}

In this section, a brief review of attention based Seq2seq model is first given. We propose modality attention mechanism in general form in subsection \ref{ssec:modalityattention} followed by integrating modality attention in Seq2Seq model in subsection \ref{ssec:modalityAttenAVSR}. Related works are discussed in subsection \ref{ssec:relatedworks}.

\subsection{End-to-end Model}
\label{ssec:end2end}

The attention based Seq2seq architecture \cite{bahdanau2014neural} consists of a sequence encoder, a sequence decoder and an attender. The encoder is usually based on an RNN which takes input as a sequence of feature vectors $\boldsymbol{x}=(x_1,x_2,\dots,x_{T})$, generating higher level representations $\textbf{h}=(h_1,h_2,\dots,h_U)$ and a final state which represents the summary of input sequence. The hidden representation of encoder is also known as memory. The final state is used to initialize the RNN decoder for predicting the output units. Because the encoding process tends to be lossy for long input sequences, an attention mechanism is introduced to automatically select the most relevant information from encoder memory to help the decoder to predict accurate unit at each decoding step.
\begin{equation}
\label{eq:listener}
\boldsymbol{h}=Encoder(\boldsymbol{x})
\end{equation}
\begin{equation}
s_i = DecoderRNN(s_{i-1},y_{i-1},c_{i-1})
\end{equation}
\begin{equation}
\label{eq:energy}
e_{i,u}=Energy(s_{i},h_u) = V^T tanh(W_h h_u+W_s s_i+b)
\end{equation}
\begin{equation}
\label{eq:wts}
\alpha_{i,u} = \frac{exp(e_{i,u})}{\sum_{u'} exp(e_{i, u'})}
\end{equation}
\begin{equation}
\label{eq:context}
c_i =\sum_{u}\alpha_{i,u}h_u
\end{equation}
\begin{equation}
\label{eq:ouputPost}
P(y_i|\boldsymbol{x},y_{<i}) = DecoderOut(s_i,c_i)
\end{equation}

Specifically, at decoding step $i$, the attender typically finishes in generating a weighted sum of the memory $\boldsymbol{h}$, which is the so called context vector $c_i$. The weights of different encoder memory are computed by their correlation score with decoder hidden state $s_i$. In the end, $c_i$ and $s_i$ is concatenated to form a context-aware state vector for the output layer to generate an output. The whole process can be summarized into Eq. (\ref{eq:listener})-(\ref{eq:ouputPost}), where $W_h$, $W_s$, $V$ and $b$ are trainable parameters. Eq. (\ref{eq:energy})-(\ref{eq:context}) tells how the attender works which is usually called content based attention.
All the parameters of this Seq2seq model are learned by minimizing the cross entropy (CE) loss  between the predict distribution and the ground truth distribution.

According to the modality of input sequence, LAS and WAS is used for auditory and visual sequence respectively. WAS usually utilizes multiple CNN layers at the bottom of encoder to extract visual features from raw pixels. In order to combine information from both audio and video, WLAS \cite{chung2017lip} uses two separate attenders to attend over the outputs of listener and watcher respectively. The acquired auditory context vector and visual context vector are then concatenated with the decoder state to give an output probability distribution.

\subsection{Modality attention mechanism}
\label{ssec:modalityattention}

The proposed modality attention mechanism fuses input from multiple modality into a single representation by weighted summing the information from individual modalities. 

Considering that there is a multimodal setup with $m=1, \dots, M$ input modalities. We suppose the input feature from each modality has 
the same length $t=1,\dots, T$ 
and is a $D$-dimensional feature vector 
$f_t^m \in {R}^D$. 
The fusion process can be summarized as follows:
\begin{equation}
\label{eq:scores}
z_{t}^m = Z(f_{1... t}^m)
\end{equation}
\begin{equation}
\label{eq:attensoftmax}
\alpha_t^m = \frac{exp(z_t^m)}{\sum_{j=1}^M exp(z_t^j)}
\end{equation}
\begin{equation}
\label{eq:weightsum}
f_t^M = \sum_{m=1}^M \alpha_t^m f_t^m
\end{equation}
Eq. (\ref{eq:scores}) represents the scoring function $Z$ which generates scores $z_t^m \in {R}^1$ based on the features of modality $m$ for time step $t$. Then a softmax operation is performed over the scores ${z_t^1, \dots, z_t^M}$ as in Eq. (\ref{eq:attensoftmax}) to get modality attention weights $\alpha_t^m \in {R}^1$. Obviously $\sum_{m=1}^M \alpha_t^m =1$. Finally in Eq. (\ref{eq:weightsum}), a weighted sum is calculated on the individual feature $f_t^m$ to obtain the fused feature representation $f_t^M $ for time step $t$.

Comparing Eq. (\ref{eq:scores})-(\ref{eq:weightsum}) with content based attention in Eq. (\ref{eq:energy})-(\ref{eq:context}), we could found they are similar in attention weights calculation and feature fusion method. At each step, the scoring function outputs a score for each modality $m$ while the energy function in  Eq. (\ref{eq:energy}) generates a score for each time step $u$. If we view features from different modalities as features from different time steps, modality attention is the similar as content based attention. The only difference maybe modality feature varies at different attention step while hidden feature $\boldsymbol{h}$ from encoder keeps constant.

The scoring function $Z$ can be any form and here we select neural networks to model it. In our experiments, we utilize a network consists of a layer of LSTM and an output layer with one node to calculate $z_t^m$:
\begin{equation}
\label{scorefunc}
Z(f_{1... t}^m) = \sigma (W \cdot LSTM(f_{1 \dots t}^m) + b)
\end{equation}
where $W$, $b$ is the parameters and $\sigma$ is the sigmoid function. Selecting LSTM as the scoring function can make it to learn the temporal variability of each modality thus make the modality attender consider past history to decide which modality is focused on more heavily at current fusion step. This is more reliable than feed forward network that only consider the current feature.

\subsection{Modality attention for AVSR}
\label{ssec:modalityAttenAVSR}
As mentioned in Section \ref{sec:intro}, AVSR is a two-modal task. Auditory and visual information is commonly concatenated for the ASR system. In AVSR, the general form of modality attention mentioned above can be used to generate a merged representation which can then be used to train a neural network. However, the frame rates of audio and video are usually different. Up sampling of video or down sampling of audio needs to perform to make sure the feature lengths are identical in order to use modality attention. We propose to use modality attention in a WLAS end-to-end system for the sake of eliminating the same feature length constraint. 

In the original WLAS architecture, after audio context vector $c_a$ and video context vector $c_v$ is computed at each decoding step, they are concatenated to the output of decoder RNN to generate an output probability distribution.  We replace the context vector concatenation process by the modality attention to calculate the merged context vector. Concretely, at decoding step $i$, $c_a$ and $c_v$ is calculated use decoder state $s_i$ with $\boldsymbol{h}_a$ and $\boldsymbol{h}_v$ which is then feed into the modality attender to obtain the modality attention weights $\alpha_i^a$ and $\alpha_i^v$. In the end $c_i^{av}$, the weighted sum of context vector $c_a$ and $c_v$ is generated to replace the concatenated context vector for computing the output probability distribution. 

By combining an additional modality attention mechanism after the attention process of WLAS framework, our system has several beneficial properties. First, it is more explicit than perform modality fusion at raw input feature level or encoder output level, since the context vectors contain information more relevant to the current output unit  after being attended by the current decoder states.  
Second, the attention weights $\alpha_i^a$ and $\alpha_i^v$ at decoding step $i$ indicate the contribution of individual modality to the merged context vector, which show model preference over modalities. 
Third, the attention weights are computed at every decoding step, their values can dynamically adjust along with the temporal changes in modality quality. This is important for the sudden or unstable noise occurs in one of the modality. By automatically adjusting attention weight to prefer the more reliable modality, modality attention in WLAS alleviate the noise effect and make AVSR more robust.
Fourth, it is a fully differentiable soft attention mechanism and suitable for end-to-end optimization, which makes attention weights to be learned automatically.
Finally, as the decode steps are much shorter than frames of raw feature or encoder output feature, it is more computationally efficient to use individual context vector to get fused feature. 

\subsection{Related works}
\label{ssec:relatedworks}
In this subsection, we compare our AVSR model to related works. As mentioned above, the main difference between our method and WLAS in \cite{chung2017lip} is we adopt an additional attention over modality context vector to select more reliable modal information while WLAS treats the two modalities equally by context vector concatenation. The AV\_Align model \cite{sterpu2018AValign} uses audio features to attend over visual features to get enhanced representation to be used for attention based decoder. The differences to their work are as follows.  First, its additional attention supplies the lip features associated with the current audio feature and occurs in the encoder side while our modality attention is integrated in decoder side which gives out the preference of modalities by attention weights. Second, ours also have the align-like process between two modal and it is performed by individual attention which generates individual context vectors associate to the current output units. Finally, denoting $T_a$, $T_v$ and $T_d$ as length of audio feature, visual feature and decoding steps respectively, it performs attention for every frame of audio representation whose complexity is $O(T_aT_v)$ while ours operates on context vector for every decoding step whose complexity is $O(2T_d)$ which is much more efficient. Our modality attention mechanism is most similar to the works in \cite{braun2018multichannelAtten, kim2017auditoryAttention} that use attention to fuse features from multi-channel audio signal. The difference lies in that we aims to combine inherently different features use attention, e.g. audio and video, while what they combine are all speech features from different microphones. While their feature length from different channel are identical,  ours has different feature length which facilitates us to operate modality attention over attended context vector along with the decoding step of an attention based decoder.

\section{experiments }
\label{sec:exps}

\subsection{Data}
\label{ssec:data}

Our audio-visual data is obtained from broadcast television news videos. It consists about 100 speakers and 104,881 examples of  speech, which is about 150 hours of data. The utterance is about 5 seconds long and contains 22 Chinese characters on average. We use another 33,026 utterances as our test set which is about 42 hours. 

The audio samples are extracted from the videos with fmpeg. We manually add white Gaussian noise to the original speech at three different SNRs, e.g. 10dB, 5dB and 0dB. Our acoustic feature is 71 dimension filter banks extracted every 10 mini-seconds within 25 mini-second window using the conventional ASR front end. First and second derivatives are not used.

The video samples have a frame speed of 25 and are preprocessed as follows. Each frame of the videos is processed with the DLib \cite{king2009dlib} face detector and the DLib face shape predictor generates 68 landmarks. Using these landmarks, we extract mouth crop from the origin images. With the mouth center calculated by landmarks of id from 49 to 68, the distance along x-axis between 12th landmark and 1st landmark is used as the mouth width $W$ and $0.8\star W$ is the mouth height $H$. 

\subsection{Baselines}
\label{ssec:baselinemodel}
As our AVSR baseline model, 4 different models are trained, namely LAS, WAS, WLAS, AV\_Align. They are all attention based end-to-end models. In order to construct truly end-to-end models, we choose Chinese character as modeling unit. The total number of modeling unit is 6812, including 26 English characters, 6784 Chinese characters, start of sequence (SOS), end of sequence (EOS) and unknown character (UNK). All models are trained by optimizing the cross entropy loss between ground-truth character transcription and predicted character sequence via adam optimizer. Curriculum learning (CL), schedule sampling (SS) and label smoothing (LS) \cite{chiu2017state-of-art-attention} are all adopted during training to improve performance. The total training epoch is set to 15. We use teacher force at the first 4 epochs and use output of last step to feed into network with schedule sampling rate that gradually increases to 0.4 from epoch 5 to epoch 8. From epoch 8 we fix schedule sampling rate to 0.4. We use an initial learning rate of 0.0002 and halve it from epoch 11. Beam search decoding is used without an external language model to evaluate our model on test sets and the beam width is set to 5. Temperature is also used in decoding to let the output distribution more smooth for better beam search results.

The encoder of LAS, is a four layer BLSTM with 256 hidden units on each LSTM. The third and fourth layers take every two consecutive frames of its input feature as input. As a result the final output representation of encoder is 4x subsampled, e.g. 25 frames per seconds. It is essential to perform down sample in encoder as our modeling units is Chinese character which is much longer than phones. The decoder is a one layer LSTM with 512 hidden nodes followed by a densely connected feed forward layer and a output softmax layer. The attention is the content based attention as in  Subsection \ref{ssec:end2end}. 

The lip regions are 3-channel RGB images resized to 64x80 pixels. 512 dimension spacial features are extracted from images by 10 layers of CNN which are then feed into a 2 layer  BLSTM to model temporal variability. Max-pooling is performed only along the width and length dimension of images. We do not perform time pooling or concatenate features in BLSTMs to reduce the time resolution. The details of watcher architecture are presented in Table \ref{tableCNNs}. The decoder and attender are the same as those of LAS. 

\begin{table}[htb]
\centering
\caption{\textit{Watcher configurations. All convolution kernels are $1X3X3$ and all maxpooling stride is $1X2X2$. Selu is a nonlinear activation in \cite{klambauer2017selu}, BN indicate batch normalization and MP represents maxpooling.}}
\begin{tabular}{|c|c|c|}
\hline
CNN layer       & operation & output size   \\ \hline \hline
0    		&   Resize    	&  Tx3x64x80\\ \hline
1-2  		&   Conv-Selu-Conv-Selu-MP-BN  & Tx32x32x40 \\ \hline
3-4  		&   Conv-Selu-Conv-Selu-MP-BN  & Tx48x16x20 \\ \hline
5-6  		&   Conv-Selu-Conv-Selu-MP-BN  & Tx72x8x10 \\ \hline
7-8  		&   Conv-Selu-Conv-Selu-MP-BN  & Tx108x4x5 \\ \hline
9-10  		&   Conv-Selu-Conv-Selu-MP-BN  & Tx128x2x2 \\ \hline
11-12		& BLSTM-BLSTM & Tx512\\ \hline
\end{tabular}
\label{tableCNNs}
\end{table}

For all our AVSR systems, we keep video signal unchanged and vary the speech signal in different SNR to explore the complementary effect of video modality for speech recognition, especially in noisy condition. WLAS model has the same encoder and decoder architecture as LAS and WAS. The only difference is that it use two attender to compute context vectors from Listener and Watcher respectively. As mentioned in \cite{chung2017lip}, WLAS is hard to train since audio signal tends to dominant in the training process. So we train Watcher and Speller first and then we fix Watcher and train Listener and Speller, in the end we train the three components jointly. AV\_Align model also use the same encoder as WLAS before the cross modality attention steps. 

We train the base models at different SNR except WAS and summarize test set CER in Table \ref{tableWERbaseline}.

\begin{table}[htb]
\centering
\caption{\textit{Recognition performance in CER [\%] of various models at different SNR.}}
\begin{tabular}{|c||c|c|c|c|}
\hline
            & 	clean & 10dB   & 5dB & 0dB\\ \hline \hline
LAS    &   7.08    &  10.33 & 12.93 & 18.65 \\ \hline
 WAS & \multicolumn{4}{c|}{44.62}   \\ \cline{1-5}
WLAS  & 7.00  &   9.07  & 10.23 &12.34 \\ \hline
AV\_align&7.6 & 10.89 & 13.69 & 19.21 \\ \hline \hline
MD\_ATT &6.95&8.54&9.87&11.93 \\ \hline
MD\_ATT\_MC &6.85&8.12&9.74&13.65 \\ \hline
\end{tabular}
\label{tableWERbaseline}
\end{table}

\subsection{Modality attention results}
\label{ssec:modalAttenResults}
We denote MD\_ATT as our proposed modality attention based AVSR system. The encoder for audio and video signal, decoder and attender of MD\_ATT are the same as WLAS.
we use 50 LSTM cells in the scoring function in Eq. (\ref{scorefunc}) for each of the modality. Four MD\_ATTs are trained with different acoustic speech and test with matched condition. We also add noise to original wav files with one of the 4 SNR randomly. Thus with this multi-condition data, we train a MD\_ATT model, denoted as MD\_ATT\_MC, and test it on different SNR test set. The two results are listed in the last two rows of Table \ref{tableWERbaseline}. 

As can be seen from Table \ref{tableWERbaseline}, although WAS performs much worse than LAS, it is helpful to integrate video information in WLAS. The contribution of video depends on the SNR of audio signal, for example, WLAS improves LAS from 18.65\% CER to 12.34\% CER in 0dB condition which is a relative 33.8\% performance gain. AV\_align model degrades in all conditions comparing to single modality LAS system. This phenomenon also happens in task LRS2 of \cite{sterpu2018AValign}.  It happens because speech can not align precisely to video when it is corrupted by noise. Our MD\_ATT model performs best among the three AVSR systems for all noisy conditions. When increasing level of noise, it shows an increased advantage and achieves a relative performance improvement up to 36\% compared to LAS system, showing the effectiveness of our multimodal attention method for audio-visual speech recognition. Trained by multi-condition speech data and video data, MD\_ATT\_MC models continue to reduce CER in all SNR except 0dB. 

We continue to report in Table \ref{tableAttWts} the average attention weights for each modality of MD\_ATT\_MC. The attention weights are computed by averaging over all decoding steps of test set, e.g. $\alpha^m=\frac{1}{N}\sum_{n=1}^N \alpha_n^m$. Since CER of WAS is much higher than that of LAS at all SNRs, we could expect our model distributes its attention more on audio than on video. Results in Table \ref{tableAttWts} shows that attention weight for audio is above 0.6 which justifies our guess. As we gradually increase noise level from clean to 0dB, MD\_ATT\_MC lowers its attention to audio from 0.641 to 0.607 and focuses more and more on video, showing an adaptation ability to speech quality. 
\begin{table}[htb]
\centering
\caption{\textit{The attention weights of MD\_ATT\_MC for audio and video modal at different test SNR averaged over all sentences in testset.}}
\begin{tabular}{|c|c|c|}
\hline
test SNR	& \multicolumn{2}{c|}{attention weights} \\ \cline{2-3}
		& $\alpha^a$	&  $\alpha^v$\\ \hline \hline
clean	& 0.641	&	0.359\\ \hline
10dB	& 0.633	&	0.367\\ \hline
5dB		& 0.624	&	0.376\\ \hline
0dB		& 0.607	& 	0.393\\ \hline
\end{tabular}
\label{tableAttWts}
\end{table}

\section{Final Remarks}
\label{sec:conclusion}
In this paper, we have proposed a multimodal attention method to fuse information from multimodal input. Modality attention mechanism is integrated in an end-to-end attention based AVSR system. Experimental results show that our proposed method obtains a 36\% relative improvement comparing to LAS in 0dB SNR and outperforms the other feature concatenation based AVSR systems. Furthermore, attention weights can automatically adjust to a more reliable modality according to the quality of individual signal. We will evaluate it with noises collected in the real field on larger AVSR dataset and  other modalities is also our next direction.

\vfill\pagebreak

\bibliographystyle{IEEEbib}
\bibliography{panzhou.2019}

\begin{thebibliography}{10}

\bibitem{chiu2017state-of-art-attention}
Chung-Cheng Chiu, Tara~N Sainath, Yonghui Wu, Rohit Prabhavalkar, Patrick
  Nguyen, Zhifeng Chen, Anjuli Kannan, Ron~J Weiss, Kanishka Rao, Katya Gonina,
  et~al.,
\newblock ``State-of-the-art speech recognition with sequence-to-sequence
  models,''
\newblock {\em arXiv preprint arXiv:1712.01769}, 2017.

\bibitem{mcgurk1976hearinglips}
Harry McGurk and John MacDonald,
\newblock ``Hearing lips and seeing voices,''
\newblock {\em Nature}, vol. 264, no. 5588, pp. 746, 1976.

\bibitem{graves2006ctc}
Alex Graves, Santiago Fern{\'a}ndez, Faustino Gomez, and J{\"u}rgen
  Schmidhuber,
\newblock ``Connectionist temporal classification: labelling unsegmented
  sequence data with recurrent neural networks,''
\newblock in {\em Proceedings of the 23rd International Conference on Machine
  Learning}. ACM, 2006, pp. 369--376.

\bibitem{graves2014ctc}
Alex Graves and Navdeep Jaitly,
\newblock ``Towards end-to-end speech recognition with recurrent neural
  networks,''
\newblock in {\em International Conference on Machine Learning}, 2014, pp.
  1764--1772.

\bibitem{graves2012rnnt}
Alex Graves,
\newblock ``Sequence transduction with recurrent neural networks,''
\newblock {\em arXiv preprint arXiv:1211.3711}, 2012.

\bibitem{graves2013ctcrnnt}
Alex Graves, Abdel-rahman Mohamed, and Geoffrey Hinton,
\newblock ``Speech recognition with deep recurrent neural networks,''
\newblock in {\em Acoustics, Speech and Signal Processing (ICASSP), 2013 IEEE
  International Conference on}. IEEE, 2013, pp. 6645--6649.

\bibitem{rao2017googleRNNT}
Kanishka Rao, Ha{\c{s}}im Sak, and Rohit Prabhavalkar,
\newblock ``Exploring architectures, data and units for streaming end-to-end
  speech recognition with rnn-transducer,''
\newblock in {\em Automatic Speech Recognition and Understanding Workshop
  (ASRU)}. IEEE, 2017, pp. 193--199.

\bibitem{bahdanau2014neural}
Dzmitry Bahdanau, Kyunghyun Cho, and Yoshua Bengio,
\newblock ``Neural machine translation by jointly learning to align and
  translate,''
\newblock {\em arXiv preprint arXiv:1409.0473}, 2014.

\bibitem{chorowski2014endFirst}
Jan Chorowski, Dzmitry Bahdanau, Kyunghyun Cho, and Yoshua Bengio,
\newblock ``End-to-end continuous speech recognition using attention-based
  recurrent nn: first results,''
\newblock {\em arXiv preprint arXiv:1412.1602}, 2014.

\bibitem{chan2016las}
William Chan, Navdeep Jaitly, Quoc Le, and Oriol Vinyals,
\newblock ``Listen, attend and spell: A neural network for large vocabulary
  conversational speech recognition,''
\newblock in {\em Acoustics, Speech and Signal Processing (ICASSP), 2016 IEEE
  International Conference on}. IEEE, 2016, pp. 4960--4964.

\bibitem{luettin1996ASM}
Juergen Luettin, Neil~A Thacker, and Steve~W Beet,
\newblock ``Visual speech recognition using active shape models and hidden
  markov models,''
\newblock in {\em ICASSP}. IEEE, 1996, pp. 817--820.

\bibitem{cootes2001AAM}
Timothy~F Cootes, Gareth~J Edwards, and Christopher~J Taylor,
\newblock ``Active appearance models,''
\newblock {\em IEEE Transactions on Pattern Analysis \& Machine Intelligence},
  , no. 6, pp. 681--685, 2001.

\bibitem{matthews2001comparison}
Iain Matthews, Gerasimos Potamianos, Chalapathy Neti, and Juergen Luettin,
\newblock ``A comparison of model and transform-based visual features for
  audio-visual lvcsr,''
\newblock in {\em IEEE International Conference on Multimedia and Expo}. IEEE,
  2001.

\bibitem{aleksic2004comparison}
Petar~S Aleksic and Aggelos~K Katsaggelos,
\newblock ``Comparison of low-and high-level visual features for audio-visual
  continuous automatic speech recognition,''
\newblock in {\em Acoustics, Speech, and Signal Processing (ICASSP). IEEE
  International Conference on}. IEEE, 2004, vol.~5, pp. V--917.

\bibitem{noda2014lipreadingCNN}
Kuniaki Noda, Yuki Yamaguchi, Kazuhiro Nakadai, Hiroshi~G Okuno, and Tetsuya
  Ogata,
\newblock ``Lipreading using convolutional neural network,''
\newblock in {\em Fifteenth Annual Conference of the International Speech
  Communication Association}, 2014.

\bibitem{tatulli2017featureCNN}
Eric Tatulli and Thomas Hueber,
\newblock ``Feature extraction using multimodal convolutional neural networks
  for visual speech recognition,''
\newblock in {\em Acoustics, Speech and Signal Processing (ICASSP), 2017 IEEE
  International Conference on}. IEEE, 2017, pp. 2971--2975.

\bibitem{chung2016lipreadWild}
Joon~Son Chung and Andrew Zisserman,
\newblock ``Lip reading in the wild,''
\newblock in {\em Asian Conference on Computer Vision}. Springer, 2016, pp.
  87--103.

\bibitem{wand2016lipreadingLSTM}
Michael Wand, Jan Koutn{\'\i}k, and J{\"u}rgen Schmidhuber,
\newblock ``Lipreading with long short-term memory,''
\newblock in {\em Acoustics, Speech and Signal Processing (ICASSP), 2016 IEEE
  International Conference on}. IEEE, 2016, pp. 6115--6119.

\bibitem{assael2016lipnet}
Yannis~M Assael, Brendan Shillingford, Shimon Whiteson, and Nando de~Freitas,
\newblock ``Lipnet: Sentence-level lipreading,''
\newblock {\em arXiv preprint}, 2016.

\bibitem{chung2017lip}
Joon~Son Chung, Andrew~W Senior, Oriol Vinyals, and Andrew Zisserman,
\newblock ``Lip reading sentences in the wild.,''
\newblock in {\em CVPR}, 2017, pp. 3444--3453.

\bibitem{petajan1984lipreading}
Eric~David Petajan,
\newblock ``Automatic lipreading to enhance speech recognition (speech
  reading),''
\newblock 1984.

\bibitem{katsaggelos2015audiovisualfusion}
Aggelos~K Katsaggelos, Sara Bahaadini, and Rafael Molina,
\newblock ``Audiovisual fusion: Challenges and new approaches,''
\newblock {\em Proceedings of the IEEE}, vol. 103, no. 9, pp. 1635--1653, 2015.

\bibitem{potamianos2003recent}
Gerasimos Potamianos, Chalapathy Neti, Guillaume Gravier, Ashutosh Garg, and
  Andrew~W Senior,
\newblock ``Recent advances in the automatic recognition of audiovisual
  speech,''
\newblock {\em Proceedings of the IEEE}, vol. 91, no. 9, pp. 1306--1326, 2003.

\bibitem{noda2015audiovisual}
Kuniaki Noda, Yuki Yamaguchi, Kazuhiro Nakadai, Hiroshi~G Okuno, and Tetsuya
  Ogata,
\newblock ``Audio-visual speech recognition using deep learning,''
\newblock {\em Applied Intelligence}, vol. 42, no. 4, pp. 722--737, 2015.

\bibitem{petridis2018end}
Stavros Petridis, Themos Stafylakis, Pingchuan Ma, Feipeng Cai, Georgios
  Tzimiropoulos, and Maja Pantic,
\newblock ``End-to-end audiovisual speech recognition,''
\newblock {\em arXiv preprint arXiv:1802.06424}, 2018.

\bibitem{sterpu2018AValign}
George Sterpu, Christian Saam, and Naomi Harte,
\newblock ``Attention-based audio-visual fusion for robust automatic speech
  recognition,''
\newblock in {\em Proceedings of the 2018 on International Conference on
  Multimodal Interaction}. ACM, 2018, pp. 111--115.

\bibitem{braun2018multichannelAtten}
Stefan Braun, Daniel Neil, Jithendar Anumula, Enea Ceolini, and Shih-Chii Liu,
\newblock ``Multi-channel attention for end-to-end speech recognition,''
\newblock {\em Proc. Interspeech 2018}, pp. 17--21, 2018.

\bibitem{kim2017auditoryAttention}
Suyoun Kim, Ian Lane, S~Kim, and I~Lane,
\newblock ``End-to-end speech recognition with auditory attention for
  multi-microphone distance speech recognition,''
\newblock {\em Proc. Interspeech 2017}, pp. 3867--3871, 2017.

\bibitem{king2009dlib}
Davis~E King,
\newblock ``Dlib-ml: A machine learning toolkit,''
\newblock {\em Journal of Machine Learning Research}, vol. 10, no. Jul, pp.
  1755--1758, 2009.

\bibitem{klambauer2017selu}
G{\"u}nter Klambauer, Thomas Unterthiner, Andreas Mayr, and Sepp Hochreiter,
\newblock ``Self-normalizing neural networks,''
\newblock in {\em Advances in Neural Information Processing Systems}, 2017, pp.
  971--980.

\end{thebibliography}

\end{document}